\pdfoutput=1
\documentclass{article}
\usepackage[dblblindworkshop,final]{neurips_2026}
\setcitestyle{numbers,square,comma}
\usepackage[utf8]{inputenc}
\usepackage[T1]{fontenc}
\usepackage[hidelinks]{hyperref}
\usepackage{url}
\usepackage{amsfonts}
\usepackage{microtype}
\usepackage{float}

\usepackage{booktabs}
\usepackage{siunitx}
\usepackage{tikz}
\usetikzlibrary{positioning}

\usepackage{soul}
\usepackage{xcolor}
\usepackage[normalem]{ulem}
\usepackage{adjustbox}
\usepackage{xurl}
\usepackage{subcaption}
\usepackage{enumitem}
\usetikzlibrary{arrows.meta,calc,positioning}

\usepackage{forest}
\useforestlibrary{edges} 

\newcommand{\ma}{\textsuperscript{*}}                 
\newcommand{\mb}{\textsuperscript{\textasciicircum}}  
\newcommand{\ab}{\textsuperscript{+}}                 

\forestset{
    EL/.style={
        edge label={node[midway, fill=white, font=\scriptsize, inner sep=2pt]{#1}}
    },
    mdatree/.style={
        forked edges,
        for tree={
            draw,
            align=center,
            font=\footnotesize,
            edge={-},
            l sep=10mm,
            s sep=8mm,
            inner sep=3pt,
        },
    },
}

\newcommand{\refstd}{rule-derived reference standard}

\newcommand{\ntrials}{$617$}

\forestset{
    mdatreec/.style={
        forked edges,
        for tree={
            draw, align=center, font=\scriptsize, edge={-},
            l sep=5mm, s sep=3mm, inner sep=2pt,
        },
    },
}

\usepackage{fancyhdr}
\newcommand{\wmhsheader}{\adjustbox{max width=\textwidth}{\footnotesize Presented as a workshop poster paper at the NeurIPS 2026 Workshop on World Models for High-Stakes Health (WMHS)}}
\fancypagestyle{plain}{%
  \fancyhf{}\fancyhead[C]{\wmhsheader}\fancyfoot[C]{\thepage}%
  }
\fancypagestyle{empty}{%
  \fancyhf{}\fancyhead[C]{\wmhsheader}%
  }
\title{TrialDesignBench: Benchmarking Language Models for Phase-I Trial Design Against Clinical Knowledge}

\workshoptitle{World Models for High-Stakes Health: Reliable Clinical Trial Simulation and Intervention-Aware Reasoning}

\author{%
  Yihan Hank Tang$^{1*}$ \quad Peng Yang$^{2*}$ \quad Jiayi Xin$^{1*}$ \quad Tianqi Shang$^{1}$ \\
  \textbf{Xuyang Chen$^{1}$ \quad Ying Yuan$^{2}$ \quad Yi Lian$^{1\dagger}$ \quad Qi Long$^{1\dagger}$} \\[4pt]
  $^{1}$University of Pennsylvania \quad
  $^{2}$The University of Texas MD Anderson Cancer Center
}

\begin{document}

\maketitle

{\renewcommand{\thefootnote}{\fnsymbol{footnote}}%
\footnotetext[1]{Equal contribution (co-first authors).}%
\footnotetext[2]{Co-senior authors.}%
\renewcommand{\thefootnote}{}%
\footnotetext{\raggedright Correspondence: Yihan Hank Tang (\texttt{hanktang@engineering.upenn.edu}), Yi Lian (\texttt{yi.lian@pennmedicine.upenn.edu}), Qi Long (\texttt{qlong@pennmedicine.upenn.edu}).}}%
\setcounter{footnote}{0}

\begin{abstract}
AI is used in much of the clinical trial pipeline. However, selecting the statistical design of a trial, a crucial step that determines how drug doses are tested on humans and evaluated, has seen almost no AI progress. LLMs appear to be a natural fit given their strengths in biomedical knowledge and statistical reasoning. Nevertheless, there is currently no appropriate way to evaluate such abilities because recent regulatory and methodological shifts, most notably FDA's Project Optimus, have redefined what constitutes a reasonable design and rendered many previous designs obsolete. To lay the groundwork for AI-assisted trial design, we introduce \textsc{TrialDesignBench}, a benchmark of 617 Phase-I/I-II oncology dose-finding trials, each paired with an expert-validated reference design that is consistent with the latest FDA regulatory initiatives. This reference standard is produced through a formalization of the expert design process. We evaluate five frontier LLMs, none of which matches the reference standard. Our ablation experiments trace this performance gap to two main causes, inference of latent, unrecorded design determinants from trial sources (supplying them raises class-wise balanced accuracy by $0.22$) and structured decision-tree reasoning ($+0.24$). \textsc{TrialDesignBench} offers a reproducible, contemporary-standard objective for evaluating and enhancing AI-assisted trial design.
\end{abstract}

\paragraph*{Data and Code Availability}
This study uses the public Aggregate Analysis of ClinicalTrials.gov (AACT)
database \citep{tasneem2012aact}.
Code and data are available at \url{https://github.com/hanktang-research/TrialDesignBench}.

\section{Introduction}
\label{sec:intro}

Clinical trials are the gold standard for validating new therapeutics \citep{friedman2015fundamentals}. However, clinical trials are costly and often fail. It is estimated that the entire development process from discovery to bringing a drug to
market costs over $\$2.5$ billion \citep{fogel2018factors}. There are generally three phases in clinical trials \citep{friedman2015fundamentals}. More than $32\%$ of trials fail at Phase~I alone \citep{wong2019estimation}. Statistical design and planning are critical components of a clinical trial, determining how doses are selected and how safety and efficacy are evaluated throughout a study \citep{friedman2015fundamentals}. The chosen statistical design can substantially affect the accuracy of these assessments and the number of patients required to reach actionable conclusions, ultimately influencing development timelines, costs, and likelihood of success. In this paper, we focus on the design method selection for phase I oncology trials, where the primary objective is to identify an appropriate dose for subsequent studies while minimizing patient exposure to subtherapeutic or overly toxic doses \citep{Yuan2024DoseOptimization}.

Although AI is applied across different parts of the clinical trial pipeline, it has made little progress in the statistical design components of clinical trials, particularly the statistical method selection. Existing AI applications in clinical trial pipelines include patient-to-trial matching \citep{jin2024matching} and trial outcome
prediction \citep{fu2022hint}. The development of these tools benefits from readily available training and benchmark data derived from historical trials, where documentations of past decisions, outcomes, or annotations can often serve as supervision signals. These signals are largely invariant to changes in regulatory guidelines.

By contrast, statistical design decisions cannot rely on historical trial designs as ground truth. Trial design selection is influenced by available statistical methods and more importantly, regulatory guidelines \citep{friedman2015fundamentals}. The design adopted in a historical study is not necessarily the design that would be recommended today. Statistical methodologies have been evolving from the traditional 3+3 design \citep{storer1989design}, to Continual Reassessment Method \citep{o1990continual}, and to the Bayesian Optimal Interval Design \citep{Yuan2016BOIN}.

Project Optimus \citep{fda2024optimus}, a landmark clinical trial design initiative published by the FDA, fundamentally breaks the traditional design paradigm. Since it is issued by the official regulatory agency, Project Optimus sets the standard for how current oncology trials are designed. It moves early-phase trials away from pursuing the maximum tolerated dose (MTD)
toward the optimal biological dose (OBD), which is to focus on both safety and efficacy rather than safety alone \citep{fourie2022optimus}. This objective defines what a reasonable trial design is. A design method well suited to identifying an MTD is not necessarily appropriate for identifying an OBD. The methodological standard has evolved alongside the objective. Regulatory recognition of contemporary Bayesian adaptive dose-finding designs was granted more recently, with BOIN receiving FDA Fit-for-Purpose designation in 2021 \citep{FDAFFP2025} and Bayesian design methods gaining broader recognition in the FDA's 2026 guidance \citep{FDA2026Bayesian}. Bayesian adaptive trial designs take advantage of the Bayes' theorem and incoming clinical data to continuously update probabilities of treatment outcomes and alter the specifics of trial implementation \citep{FDA2026Bayesian}.
Since both the design objective and the accepted methodology have changed fundamentally, the overwhelming majority of recorded trials can no longer serve as ground truth or training data for models expected to meet the latest clinical trial guidelines.

We introduce \textsc{TrialDesignBench}, the \textbf{first} benchmark for AI-assisted statistical design recommendation in clinical trials. Our contributions are as follows:
\begin{enumerate}[nosep,leftmargin=*]
\item  We curate a benchmark of \textbf{617} early-phase oncology dose-finding trials, each annotated with rich clinical information fields and paired with an expert-validated reference-standard design over \textbf{11} trial design method classes. The reference standard is produced through a formalization of the expert design process and reflects Project Optimus guidelines.
    \item We evaluate \textbf{five} frontier LLMs under tiered input information configurations that isolate the effects of clinical information availability and expert-derived design principles. No model matches the reference standard, demonstrating a substantial performance gap.
    \item Through ablations, we identify two consistent bottlenecks in LLM-based design recommendation, recovering unrecorded, latent design information from source text ($+0.22$ class-wise balanced accuracy when supplied) and structured decision-tree
    reasoning ($+0.24$).
\end{enumerate}

Together, \textsc{TrialDesignBench} establishes a reproducible, modern-standard foundation for robust evaluation and future improvements in AI systems for statistical trial design.

\section{Benchmark Construction}
\label{sec:corpus}

\subsection{Data Collection}
\label{sec:corpus:frame}
\begin{figure}[t]
\centering
    \includegraphics[width=0.95\textwidth]{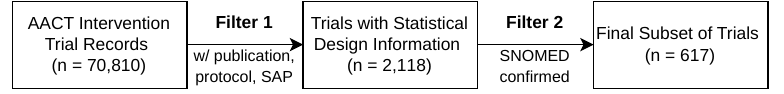}
\caption{Corpus selection process.}
\vspace{6pt}
\label{fig:funnel}
\end{figure}

We draw our sampling frame from the AACT database \citep{tasneem2012aact},
comprising approximately 70{,}810 interventional  trial records (fetched on 28 May 2026). We
restrict the selection to Phase~I and combined Phase~I \& II dose-finding interventional oncology trials with a linked publication on PubMed and a posted protocol and, where available, a statistical analysis plan (SAP). The registry reports information such as trial phase,
conditions, and interventions. Protocols and SAPs contain more crucial information, such as the dose-limiting toxicity (DLT) definition, the DLT assessment window, and the endpoints used to select the dose, for biostatisticians to perform trial design recommendation
(Section~\ref{sec:results:latents}). As illustrated in Figure~\ref{fig:funnel}, after applying
these filters, we derive \ntrials{} trials.

\subsection{Benchmark Inputs}
\label{sec:corpus:taxonomy}

For each trial, we curate three levels of trial attributes that reflect progressively richer information available during statistical design selection, enabling controlled evaluation of LLMs' ability to recommend design methods by varying the amount of trial knowledge provided as input.

\paragraph{Basic Trial Context} High-level trial information, including condition, intervention, phase, and objectives. These are typically communicated to biostatisticians during initial discussions with physician trialists. For benchmark curation, Basic Trial Context is \textbf{extracted} from ClinicalTrials.gov records.

\paragraph{Trial Attributes} More fine-grained trial information relevant to operational and planning details, including endpoint definition, treatment schedules, and other design-related characteristics enumerated in Appendix~\ref{app:conditions}. These details are typically documented during the very early phases of protocol development and before final decision on any statistical design selection. For benchmark curation, Trial Attributes are \textbf{extracted} from protocols and SAPs.

\paragraph{Design Factors} Design-relevant attributes that are not explicitly described in protocols and SAPs but can be \textbf{inferred}. Examples include treatment regimen type (single agent vs. combination), inferred accrual rate, primary endpoint type (acute vs.\ delayed) \citep{Chen2025TITEBOIN}, and others (Appendix~\ref{app:conditions}). These factors approximate information that trialists and biostatisticians often derive from domain knowledge when evaluating candidate statistical designs. For benchmark curation, such factors were inferred and validated based on basic trial context and trial attributes.

\subsection{Reference-standard Label}
\label{sec:corpus:refstd} 
\begin{figure}[t]
  \centering
  \includegraphics[width=\linewidth]{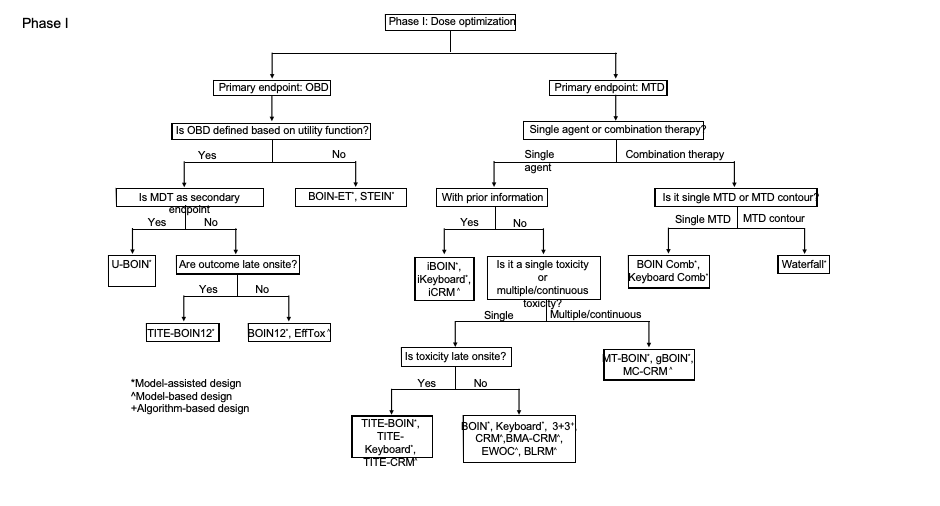}
  \caption{Decision workflow for Phase~I dose-finding interventional oncology trial
  design (compact overview). A full-size, legible version, split by primary endpoint
  into the MTD and OBD branches, is provided in Figure~\ref{fig:mdatree-full}
  (Appendix~\ref{apd:rules}).}
  \label{fig:mdatree-1}
\end{figure}

The reference-standard recommendation for each trial is produced based on contemporary phase~I dose-finding design principles developed and adopted by experienced trial methodologists and biostatisticians. Figure~\ref{fig:mdatree-1} summarizes this workflow as a decision tree adapted from modern dose-finding methodology \citep{Yuan2022ModelAssisted,Yuan2024DoseOptimization}. The tree sequentially evaluates key Basic Trial Context, Trial Attributes, and Design Factors, a subset of the benchmark inputs described in Section~\ref{sec:corpus:taxonomy}, ultimately leading to a recommended design method.

The benchmark output is a single reference-standard design recommendation. Each recommendation corresponds to a specific dose-finding method, such as BOIN, 3+3, or CRM. Although a single method is assigned to each trial, multiple methods may represent reasonable choices to the same dose-finding problem. Methods grouped in the same leaf node of the decision workflow (Figure~\ref{fig:mdatree-1}) are grouped into a class, reflecting shared design objectives and methodological characteristics.

The resulting reference-standard recommendations were subsequently reviewed and validated by experienced trial methodologists and biostatisticians to ensure consistency with contemporary methodological standards rather than historical practice. The node-resolution rules, in particular the late-onset determination (Appendix~\ref{apd:lot}), were calibrated against case sets labeled by these experts, and the experts re-verified the single-agent-versus-combination node for every trial initially labeled with a combination design, which reassigned 31 of 51 such trials to single-agent designs. The complete decision rules are provided in Appendix~\ref{apd:rules}, and the label distribution is reported in Appendix~\ref{app:labeldist}.

\section{Experiments}
We evaluate frontier LLMs on \textsc{TrialDesignBench} to establish baseline performance for AI-assisted statistical trial design recommendation. In addition, we investigate how recommendation quality changes as progressively richer trial information becomes available and whether expert-guided decision-tree reasoning improves performance. Our experiments aim to (1) quantify the current capabilities of LLMs on statistical trial design recommendation, (2) assess whether LLM performance improves with increasingly detailed trial information and access to expert-derived design principles encoded in Figure~\ref{fig:mdatree-1}, and (3) identify common failure modes in model reasoning.

\subsection{Evaluation Metric}
\label{sec:eval}

Each model recommendation and each reference-standard label correspond to a specific design method. Since multiple design methods may be admissible for the same trial, we intentionally relax evaluation from exact reference-standard matching to class-level agreement. Let $c(\cdot)$ denote the mapping from a design method to its class. A recommendation $m$ is considered correct whenever $c(m)=c(m^\star)$, where $m^\star$ is the reference-standard recommendation. For $N$ evaluated trials with recommendations $m_i$ and reference-standard recommendations $m_i^\star$, class-wise accuracy is then
\begin{equation}
\mathrm{Acc}_c=\frac{1}{N}\sum_{i=1}^{N}\mathbb{I}\left[c(m_i)=c(m_i^\star)\right].
\end{equation} We report it together with class-wise balanced accuracy, macro-F1, and Cohen's~$\kappa$ (Section~\ref{sec:results}) throughout. The complete class taxonomy is provided in Appendix~\ref{app:classes}.

\subsection{Model Selection}

To comprehensively evaluate the performance of popular LLMs on trial design recommendation, we select Claude Opus 5, GPT-5, OpenAI o3, OpenAI gpt-oss-120b, and Llama 4 Maverick.

\subsection{Input Configurations}
\label{sec:ablation}
To understand the information and reasoning requirements of statistical trial design recommendation, we construct a series of input configurations that progressively enrich the trial and design information available to the model. The configurations form a tiered information hierarchy crossed with an expert-derived design workflow:

\begin{description}

\item[\textbf{A. Basic Trial Context.}] Basic Trial Context only.

\item[\textbf{B. A + Trial Attributes.}] Basic Trial Context and Trial Attributes.

\item[\textbf{C. B + Design Factors.}] Basic Trial Context, Trial Attributes, and Design Factors.

\item[\textbf{D. B + Workflow.}] Basic Trial Context, Trial Attributes, and the decision workflow (Figure~\ref{fig:mdatree-1}).

\item[\textbf{E. C + Workflow.}] Basic Trial Context, Trial Attributes, Design Factors, and the decision workflow.

\end{description}

Configurations A$\rightarrow$B$\rightarrow$C evaluate whether recommendation performance improves as progressively richer trial information, particularly design factors that required judgment by experienced trialists, becomes available. Configurations D and E additionally provide the design workflow, allowing the contributions of trial information and design principles to be examined separately. Because configuration E supplies both the design factors and the decision workflow from which the reference labels are derived, it can be viewed as an oracle-workflow condition.

\subsection{Implementation Details}

All five models were accessed through a single unified serving layer
with pinned versions and
identical prompting and decoding settings, so that differences across models
reflect capability. For each of the 617 trials under input configurations A, B, and C, every model
returns a direct design class  recommendation. For input configurations D and E, every model returns a structured JSON response, per-node binary answers with rationales, which a deterministic, code-based tree walker maps to a single leaf
design class. Predictions are then scored against the expert validated labels using the
class-level accuracy defined in Section~\ref{sec:eval}. For configuration E, the design-factor input at the combination-target node was corrected after review to match the field used by the reference rules, and the 20 affected combination trials were re-run. All E results reflect this correction.

\section{Results}
\label{sec:results}
We report four class-wise metrics, accuracy ($\mathrm{Acc}_c$), balanced
accuracy ($\mathrm{BalAcc}_c$), macro-F1 ($\mathrm{F1}_c$), and Cohen's~$\kappa$ ($\kappa_c$), each computed after mapping the
recommended and validated methods to their design-method classes
(Section~\ref{sec:eval}). Balanced accuracy is the mean per-class recall, macro-F1 the unweighted mean of per-class F1, and $\kappa$ Cohen's chance-corrected agreement. The subscript $c$ marks computation over design-method classes rather than raw methods.
Table~\ref{tab:main} summarizes performance averaged over the five
models. Table~\ref{tab:balacc} gives the per-model balanced accuracy underlying
the two findings below.

\begin{table}[t]
\centering\small
\setlength{\tabcolsep}{4pt}
\caption{Class-wise metrics averaged over five models, by input configuration.
Input configurations form an information ladder (A$\to$B$\to$C) crossed with a
reasoning scaffold (B$\to$D, C$\to$E).}
\vspace{6pt}
\label{tab:main}
\adjustbox{max width=\linewidth}{%
\begin{tabular}{lcccc}
\toprule
Input Configuration & Acc$_{c}$ & BalAcc$_{c}$ & F1$_{c}$ & $\kappa_{c}$ \\
\midrule
A \,(basic trial context)      & 0.436 & 0.225 & 0.099 & 0.059 \\
B \,(A + attributes)       & 0.614 & 0.307 & 0.135 & 0.131 \\
C \,(B + design factors)      & 0.806 & 0.531 & 0.293 & 0.406 \\
D \,(B + decision workflow)      & 0.715 & 0.547 & 0.298 & 0.272 \\
E \,(C + decision workflow)      & \textbf{0.868} & \textbf{0.728} & \textbf{0.506} & \textbf{0.542} \\
\bottomrule
\end{tabular}}

\end{table}

\begin{table}[t]
\centering\small
\caption{Class-wise \emph{balanced accuracy} per model. Both effects are
5/5 consistent (C$>$B and D$>$B on all four metrics; Appendix~\ref{app:counts}).}
\vspace{6pt}
\label{tab:balacc}
\setlength{\tabcolsep}{4pt}
\adjustbox{max width=\linewidth}{%
\begin{tabular}{lccccc}
\toprule
Model & A & B & C & D & E \\
\midrule
Claude Opus 5      & 0.152 & 0.395 & 0.558 & 0.499 & 0.764 \\
GPT-5              & 0.259 & 0.327 & 0.662 & 0.570 & 0.787 \\
o3                 & 0.219 & 0.314 & 0.408 & 0.552 & 0.761 \\
gpt-oss-120b       & 0.276 & 0.254 & 0.508 & 0.450 & 0.480 \\
Llama 4 Maverick   & 0.219 & 0.246 & 0.522 & 0.663 & 0.850 \\
\midrule
Average & 0.225 & 0.307 & 0.531 & 0.547 & 0.728 \\
\bottomrule
\end{tabular}}

\end{table}

\subsection{Model Performance Degrades Without Inferred Design Factors}
\label{sec:results:latents}
Supplying the inferred design factors (input configuration C vs. B) improves
every class-wise metric for all five models. Averaged over models,
class-wise accuracy rises by $+0.192$ ($0.614\!\to\!0.806$), balanced accuracy
by $+0.224$ ($0.307\!\to\!0.531$), macro-F1 by $+0.158$ ($0.135\!\to\!0.293)$, and $\kappa$ by
$+0.275$ ($0.131\!\to\!0.406$). The same performance gap exists under the reasoning
scaffold ($E$ vs.\ $D$, $+0.153$ class-wise accuracy). Therefore, the bottleneck in frontier LLM recommendation appears to be the translation of trial information into latent design factors that influence design selection. For example, when a trial combines an investigational agent with a standard-of-care therapy of known dose and toxicity profile, models may incorrectly formulate the problem as a two-dimensional dose-finding study and recommend a combination design.

A node-level breakdown of the tree walks (Appendix~\ref{app:counts}, Table~\ref{tab:nodedev}) localizes where this inference fails. Deviations from the reference concentrate at the late-onset node, the accrual-to-window latent, which alone accounts for more incorrect node answers than all other nodes combined, followed by the single-versus-combination node. The single-versus-combination node shows a deceptively low deviation rate. This is because the dataset is dominated by the single-agent class trials. True-Combination recall at this node ranges from $20\%$ to $45\%$ across models.

\subsection{Structured Reasoning Recovers Minority-Class Performance}
\label{sec:results:scaffold}
Requiring the model to traverse the expert-codified decision tree, answering
each node with a binary decision and rationale
rather than naming a design directly, improves performance.
Adding the reasoning scaffold to the structured inputs ($D$ vs.\ $B$) raises
model-averaged balanced accuracy by $+0.240$ ($0.307\!\to\!0.547$), macro-F1 by
$+0.163$, $\kappa$ by $+0.141$, and class-wise accuracy by $+0.101$. These four improvements are consistent across models. The benefit is concentrated in the imbalance-sensitive
metrics. On balanced accuracy, the scaffold \emph{alone}
(D, $0.547$) matches the hand-resolved-latent input configurations (C, $0.531$), which means that
structured reasoning is capable of making up for most of the value of the expert-supplied latents.
When those latents are already given, the additional scaffold gain shrinks
($E$ vs.\ $C$: $+0.197$ balanced accuracy, positive for 4/5 models). Structured reasoning, through serializing and visualizing the structured reasoning process, makes it straightforward to identify reasoning errors and implement fixes.

\section{Limitations and Future Work}
\label{sec:limitations}

Our benchmark has several limitations. First, we recommend a single reference-standard trial design when multiple design methods are often admissible. We address this by grouping multiple design methods into the same class in the workflow and running experiments within defined design-method classes. Second, to construct the first benchmark dataset, we restrict the scope to a more homogeneous set of Phase~I/I--II oncology dose-finding trials to ensure high-quality extraction of trial information and design method recommendations. For the same purpose, we select trials with a linked publication and a posted protocol so that trial information can be more accurately derived without significant loss of representativeness or generalizability for the purpose of benchmarking.

We will expand to a broader range of phase I--III trials, with or without protocol documents. A major challenge is that much design-relevant information is not explicitly reported in clinicaltrials.gov registry records. The current \textsc{TrialDesignBench} may serve as few-shot examples for automated inference of latent design factors, facilitating scalable benchmark development.

\section{Conclusion}
\label{sec:conclusion}
We introduce \textsc{TrialDesignBench}, a benchmark of \ntrials{} Phase~I and
Phase~I/II oncology dose-finding trials paired with a \refstd{} for the
recommended statistical design. The workflow that produces the benchmark is based on contemporary dose-finding methodology
rather than historically adopted practice. The benchmark reframes trial-design
recommendation as an evaluable task. It enables us to measure a suite of contemporary LLMs against it under common
class-wise criteria.

By modeling the trial design recommendation process as sequential and having models decide on nodes before traversing the decision tree, human experts are able to localize where recommendations break down. Models are
limited considerably by inferring the latent,
structurally unrecorded fields of trial design such as regimen modality
and the late-onset accrual-to-window ratio.
Without those resolved fields, models perform considerably worse on every metric. In addition, models that do not follow the tree-guided,
expert-codified reasoning scaffold also show a considerable drop in performance. No model matches the rule-derived reference standard,
leaving clear headroom.

Beyond ranking today's systems, \textsc{TrialDesignBench} provides a reusable,
Project-Optimus-aligned target for AI-assisted trial design and a diagnostic that
separates extraction from higher-level inference. We see it as a step toward
tools that help investigators move from basic trial information to a
statistically sound design, and ultimately to trial protocol development.

\section*{Responsible-Use Statement}
\label{sec:responsible}

\textsc{TrialDesignBench} is an evaluation benchmark, not a clinical decision tool.
Its reference-standard designs are a rule-based formalization of one specified expert workflow for contemporary Phase-I
dose-finding methodology. Although it is expert-validated, it is not a unique clinical gold standard and not a substitute for a qualified
biostatistician. No results here should inform a real trial without expert review.
Every evaluated model in this paper fails to match the reference standard and can be confidently
wrong. Therefore, unsupervised use risks propagating unsafe or inefficient designs. The
corpus is built from public AACT/ClinicalTrials.gov and PubMed-linked protocols and
may inherit their selection and documentation biases. The corpus largely consists of completed and well-documented oncology
trials. We release processed data to facilitate future research on AI for clinical trial design. We recommend a domain
expert in the loop with calibrated uncertainty for any downstream use.

\bibliographystyle{plainnat}

\bibliography{ref}

\begin{thebibliography}{16}
\providecommand{\natexlab}[1]{#1}
\providecommand{\url}[1]{\texttt{#1}}
\expandafter\ifx\csname urlstyle\endcsname\relax
  \providecommand{\doi}[1]{doi: #1}\else
  \providecommand{\doi}{doi: \begingroup \urlstyle{rm}\Url}\fi

\bibitem[Chen et~al.(2025)Chen, Chen, Zhang, Lin, and Yuan]{Chen2025TITEBOIN}
Kai Chen, Ting-Yu Chen, Yiming Zhang, Ruitao Lin, and Ying Yuan.
\newblock Practical considerations for using the {TITE-BOIN} design to handle late-onset toxicity or fast accrual in phase {I} trials.
\newblock \emph{Clinical Cancer Research}, 31\penalty0 (13):\penalty0 2573--2580, 2025.

\bibitem[Fogel(2018)]{fogel2018factors}
David~B Fogel.
\newblock Factors associated with clinical trials that fail and opportunities for improving the likelihood of success: a review.
\newblock \emph{Contemporary clinical trials communications}, 11:\penalty0 156--164, 2018.

\bibitem[Fourie~Zirkelbach et~al.(2022)Fourie~Zirkelbach, Shah, Vallejo, Cheng, Ayyoub, Liu, Hudson, Sridhara, Ison, Amiri-Kordestani, et~al.]{fourie2022optimus}
Jeanne Fourie~Zirkelbach, Mirat Shah, Jonathon Vallejo, Joyce Cheng, Amal Ayyoub, Jiang Liu, Rachel Hudson, Rajeshwari Sridhara, Gwynn Ison, Laleh Amiri-Kordestani, et~al.
\newblock Improving dose-optimization processes used in oncology drug development to minimize toxicity and maximize benefit to patients.
\newblock \emph{Journal of Clinical Oncology}, 40\penalty0 (30):\penalty0 3489--3500, 2022.

\bibitem[Friedman et~al.(2015)Friedman, Furberg, DeMets, Reboussin, and Granger]{friedman2015fundamentals}
Lawrence~M. Friedman, Curt~D. Furberg, David~L. DeMets, David~M. Reboussin, and Christopher~B. Granger.
\newblock \emph{Fundamentals of Clinical Trials}.
\newblock Springer, Cham, 5th edition, 2015.

\bibitem[Fu et~al.(2022)Fu, Huang, Xiao, Glass, and Sun]{fu2022hint}
Tianfan Fu, Kexin Huang, Cao Xiao, Lucas~M. Glass, and Jimeng Sun.
\newblock {HINT}: Hierarchical interaction network for clinical-trial-outcome predictions.
\newblock \emph{Patterns}, 3\penalty0 (4):\penalty0 100445, 2022.
\newblock \doi{10.1016/j.patter.2022.100445}.

\bibitem[Jin et~al.(2024)Jin, Wang, Floudas, Chen, Gong, Bracken-Clarke, Xue, Yang, Sun, and Lu]{jin2024matching}
Qiao Jin, Zifeng Wang, Charalampos~S Floudas, Fangyuan Chen, Changlin Gong, Dara Bracken-Clarke, Elisabetta Xue, Yifan Yang, Jimeng Sun, and Zhiyong Lu.
\newblock Matching patients to clinical trials with large language models.
\newblock \emph{Nature communications}, 15\penalty0 (1):\penalty0 9074, 2024.

\bibitem[O'Quigley et~al.(1990)O'Quigley, Pepe, and Fisher]{o1990continual}
John O'Quigley, Margaret Pepe, and Lloyd Fisher.
\newblock Continual reassessment method: a practical design for phase 1 clinical trials in cancer.
\newblock \emph{Biometrics}, pages 33--48, 1990.

\bibitem[Storer(1989)]{storer1989design}
Barry~E Storer.
\newblock Design and analysis of phase {I} clinical trials.
\newblock \emph{Biometrics}, pages 925--937, 1989.

\bibitem[Tasneem et~al.(2012)Tasneem, Aberle, Ananth, Chakraborty, Chiswell, McCourt, and Pietrobon]{tasneem2012aact}
Asba Tasneem, Laura Aberle, Hari Ananth, Swati Chakraborty, Karen Chiswell, Brian~J McCourt, and Ricardo Pietrobon.
\newblock The database for aggregate analysis of clinicaltrials.gov ({AACT}) and subsequent regrouping by clinical specialty.
\newblock \emph{PloS one}, 7\penalty0 (3):\penalty0 e33677, 2012.

\bibitem[{U.S. Food and Drug Administration}(2024)]{fda2024optimus}
{U.S. Food and Drug Administration}.
\newblock Project {Optimus}: Reforming the dose optimization and dose selection paradigm in oncology.
\newblock \url{https://www.fda.gov/about-fda/oncology-center-excellence/project-optimus}, 2024.
\newblock Oncology Center of Excellence. Accessed September 10, 2026.

\bibitem[{U.S. Food and Drug Administration}(2025)]{FDAFFP2025}
{U.S. Food and Drug Administration}.
\newblock Drug development tools: Fit-for-purpose initiative, 2025.
\newblock URL \url{https://www.fda.gov/drugs/development-approval-process-drugs/drug-development-tools-fit-purpose-initiative}.
\newblock Accessed September 10, 2026.

\bibitem[{U.S. Food and Drug Administration}(2026)]{FDA2026Bayesian}
{U.S. Food and Drug Administration}.
\newblock Use of {B}ayesian methodology in clinical trials of drug and biological products, 2026.
\newblock URL \url{https://www.fda.gov/regulatory-information/search-fda-guidance-documents/use-bayesian-methodology-clinical-trials-drug-and-biological-products}.
\newblock Accessed September 10, 2026.

\bibitem[Wong et~al.(2019)Wong, Siah, and Lo]{wong2019estimation}
Chi~Heem Wong, Kien~Wei Siah, and Andrew~W. Lo.
\newblock Estimation of clinical trial success rates and related parameters.
\newblock \emph{Biostatistics}, 20\penalty0 (2):\penalty0 273--286, 2019.

\bibitem[Yuan et~al.(2016)Yuan, Hess, Hilsenbeck, and Gilbert]{Yuan2016BOIN}
Ying Yuan, Kenneth~R. Hess, Susan~G. Hilsenbeck, and Mark~R. Gilbert.
\newblock {B}ayesian optimal interval design: A simple and well-performing design for phase {I} oncology trials.
\newblock \emph{Clinical Cancer Research}, 22\penalty0 (17):\penalty0 4291--4301, 2016.

\bibitem[Yuan et~al.(2022)Yuan, Lin, and Lee]{Yuan2022ModelAssisted}
Ying Yuan, Ruitao Lin, and J.~Jack Lee.
\newblock \emph{Model-Assisted {B}ayesian Designs for Dose Finding and Optimization: Methods and Applications}.
\newblock CRC Press, 2022.

\bibitem[Yuan et~al.(2024)Yuan, Zhou, and Liu]{Yuan2024DoseOptimization}
Ying Yuan, Heng Zhou, and Suyu Liu.
\newblock Statistical and practical considerations in planning and conduct of dose-optimization trials.
\newblock \emph{Clinical Trials}, 21\penalty0 (3):\penalty0 273--286, 2024.

\end{thebibliography}

\clearpage
\appendix
\section*{Appendix}

\section{Design-Method Class Taxonomy}
\label{app:classes}
Recommendations and reference-standard labels are grouped into eleven classes,
the leaves of the decision tree (Figure~\ref{fig:mdatree-full}). Class-wise
credit ($c(m)=c(m^\star)$, Section~\ref{sec:eval}) is given within
Classes~1--10; Class~11 is a residual catch-all of off-tree designs and receives
credit only on an exact-method match.
\begin{enumerate}\setlength{\itemsep}{1pt}
  \item[] \textbf{1. Standard single-agent MTD:} BOIN, Keyboard, 3+3, CRM,
  BMA-CRM, EWOC, BLRM.
  \item[] \textbf{2. Time-to-event single-agent MTD:} TITE-BOIN, TITE-Keyboard,
  TITE-CRM.
  \item[] \textbf{3. Multiple
  / Continuous toxicity:} MT-BOIN, gBOIN, MC-CRM.
  \item[] \textbf{4. Informative-prior:} iBOIN, iKeyboard, iCRM.
  \item[] \textbf{5. Combination single-MTD:} BOIN Comb, Keyboard Comb.
  \item[] \textbf{6. Combination MTD-contour:} Waterfall.
  \item[] \textbf{7. OBD time-to-event:} TITE-BOIN12.
  \item[] \textbf{8. OBD without utility:} BOIN-ET, STEIN.
  \item[] \textbf{9. OBD utility-based:} BOIN12, EffTox.
  \item[] \textbf{10. OBD with MTD as secondary:} U-BOIN.
  \item[] \textbf{11. Other (off-tree):} TPI, mTPI, mTPI-2, TEQR, Rolling-6,
  Storer two-stage, accelerated titration.
\end{enumerate}

\section{Reference-Standard Label Distribution}
\label{app:labeldist}
Table~\ref{tab:labeldist} reports the design-method label distribution of the
reference standard across the \ntrials{} trials.

\begin{table}[H]
\centering\small
\caption{Reference-standard label distribution (\ntrials{} trials).}
\vspace{6pt}
\label{tab:labeldist}
\begin{tabular}{lrr}
\toprule
Label & Count & \%\\
\midrule
BOIN         & 366 & 59.3\\
Out of scope & 142 & 23.0\\
3+3          & 59  & 9.6\\
TITE-BOIN    & 24  & 3.9\\
BOIN Comb    & 20  & 3.2\\
BOIN-ET      & 4   & 0.6\\
TITE-BOIN12  & 1   & 0.2\\
BOIN12       & 1   & 0.2\\
\midrule
Total        & 617 & 100.0\\
\bottomrule
\end{tabular}

\end{table}

Within the standard single-agent MTD class (Class~1, Appendix~\ref{app:classes}), trials were initially labelled \textbf{BOIN}. However, a trial is relabeled \textbf{3+3} when it prespecifies at most two
dose levels.
 Two things are worth noting here:
\begin{itemize}\setlength{\itemsep}{1pt}
  \item The fact that 3+3 methods are recommended here is primarily grounded in operational simplicity.
  \item The prevalence of BOIN and 3+3 over other design methods reflects statistical soundness and operational consideration.
\end{itemize}

\section{Input Configurations}
\label{app:conditions}
Every input configuration identifies the trial by a case ID only. The originally adopted
design, reported results, and design-revealing fields are withheld
(Appendix~\ref{apd:prompts}).

\paragraph{A (protocol narrative).} The protocol's background, rationale, and
objectives excerpt only. No structured fields.

\paragraph{B (structured fields).} Fourteen fields: brief title, phase,
conditions, interventions, enrollment, start date, Phase~I objective, endpoint
nature, DLT (Dose-limiting Toxicity) definition, CTCAE (Common Terminology Criteria for Adverse Events) version, DLT assessment window, interpatient
interval, interval basis, and safety-stagger floor, together with protocol excerpts in A.

\paragraph{C (structured + resolved latents).} All of~B, along with seven resolved
latent fields, primary endpoint type, toxicity multiplicity, prior-information
availability, regimen modality, combination target, OBD basis, MTD-as-secondary, and the resolved late-onset quantities (the accrual-to-window
ratio $W/I$, the escalation gate, and the interval basis).

\paragraph{D, E.} Identical field inputs to B and C respectively, presented under the
tree-walk prompt (Appendix~\ref{apd:prompts}).

\section{Full Per-Model Results and Scoring Counts}
\label{app:counts}

Table~\ref{tab:fullmetrics} reports every model's class-wise accuracy,
macro-F1, and Cohen's~$\kappa$ across the five input configurations (balanced accuracy is
in Table~\ref{tab:balacc}). It substantiates the 5/5 consistency of C$>$B and
D$>$B stated in the main text.

\begin{table}[H]
\centering\small
\setlength{\tabcolsep}{4pt}
\caption{Per-model class-wise metrics by input configuration (five models).}
\vspace{6pt}
\label{tab:fullmetrics}
\adjustbox{max width=\linewidth}{%
\begin{tabular}{lccccc}
\toprule
Model & A & B & C & D & E \\
\midrule
\multicolumn{6}{l}{\emph{Class-wise accuracy}}\\
Claude Opus 5     & 0.248 & 0.598 & 0.907 & 0.838 & 0.955 \\
GPT-5             & 0.264 & 0.552 & 0.860 & 0.639 & 0.895 \\
o3                & 0.436 & 0.601 & 0.725 & 0.653 & 0.794 \\
gpt-oss-120b      & 0.482 & 0.554 & 0.720 & 0.620 & 0.829 \\
Llama 4 Maverick  & 0.749 & 0.767 & 0.818 & 0.823 & 0.864 \\
\textit{Mean}     & 0.436 & 0.614 & 0.806 & 0.715 & 0.868 \\
\midrule
\multicolumn{6}{l}{\emph{Class-wise macro-F1}}\\
Claude Opus 5     & 0.113 & 0.135 & 0.397 & 0.273 & 0.562 \\
GPT-5             & 0.064 & 0.119 & 0.349 & 0.355 & 0.620 \\
o3                & 0.108 & 0.175 & 0.231 & 0.384 & 0.597 \\
gpt-oss-120b      & 0.087 & 0.095 & 0.153 & 0.223 & 0.313 \\
Llama 4 Maverick  & 0.121 & 0.150 & 0.332 & 0.252 & 0.440 \\
\textit{Mean}     & 0.099 & 0.135 & 0.293 & 0.298 & 0.506 \\
\midrule
\multicolumn{6}{l}{\emph{Class-wise $\kappa$}}\\
Claude Opus 5     & 0.040 & 0.140 & 0.586 & 0.382 & 0.778 \\
GPT-5             & 0.045 & 0.141 & 0.518 & 0.204 & 0.598 \\
o3                & 0.037 & 0.138 & 0.294 & 0.193 & 0.435 \\
gpt-oss-120b      & 0.083 & 0.097 & 0.270 & 0.188 & 0.392 \\
Llama 4 Maverick  & 0.087 & 0.142 & 0.361 & 0.392 & 0.506 \\
\textit{Mean}     & 0.059 & 0.131 & 0.406 & 0.272 & 0.542 \\
\bottomrule
\end{tabular}}

\end{table}

\begin{table}[H]
\centering\small
\setlength{\tabcolsep}{6pt}
\caption{Node-level deviation rate for the tree-walk configurations (D and~E),
averaged over the five models: the fraction of trials whose model node answer
disagrees with the reference at each node. }
\vspace{6pt}
\label{tab:nodedev}
\begin{tabular}{lccr}
\toprule
Node & D & E & Pooled $n$\\
\midrule
\texttt{mtd\_late\_onset}   & 0.262 & 0.118 & 3{,}919\\
\texttt{mtd\_comb\_target}  & 0.011 & 0.000 & 172\\
\texttt{mtd\_modality}      & 0.029 & 0.027 & 4{,}627\\
\bottomrule
\end{tabular}

\end{table}

\section{Decision Rules}
\label{apd:rules}
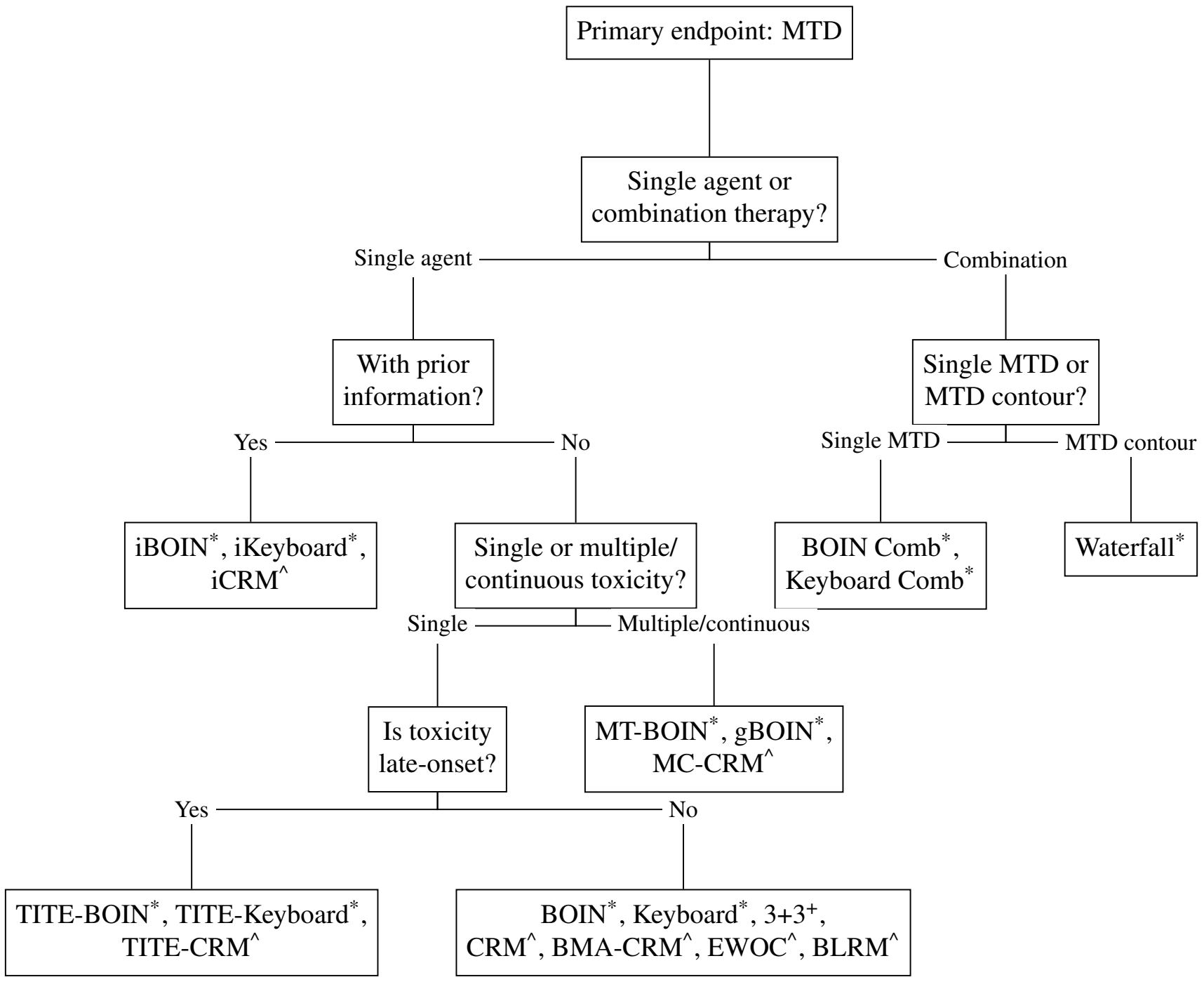
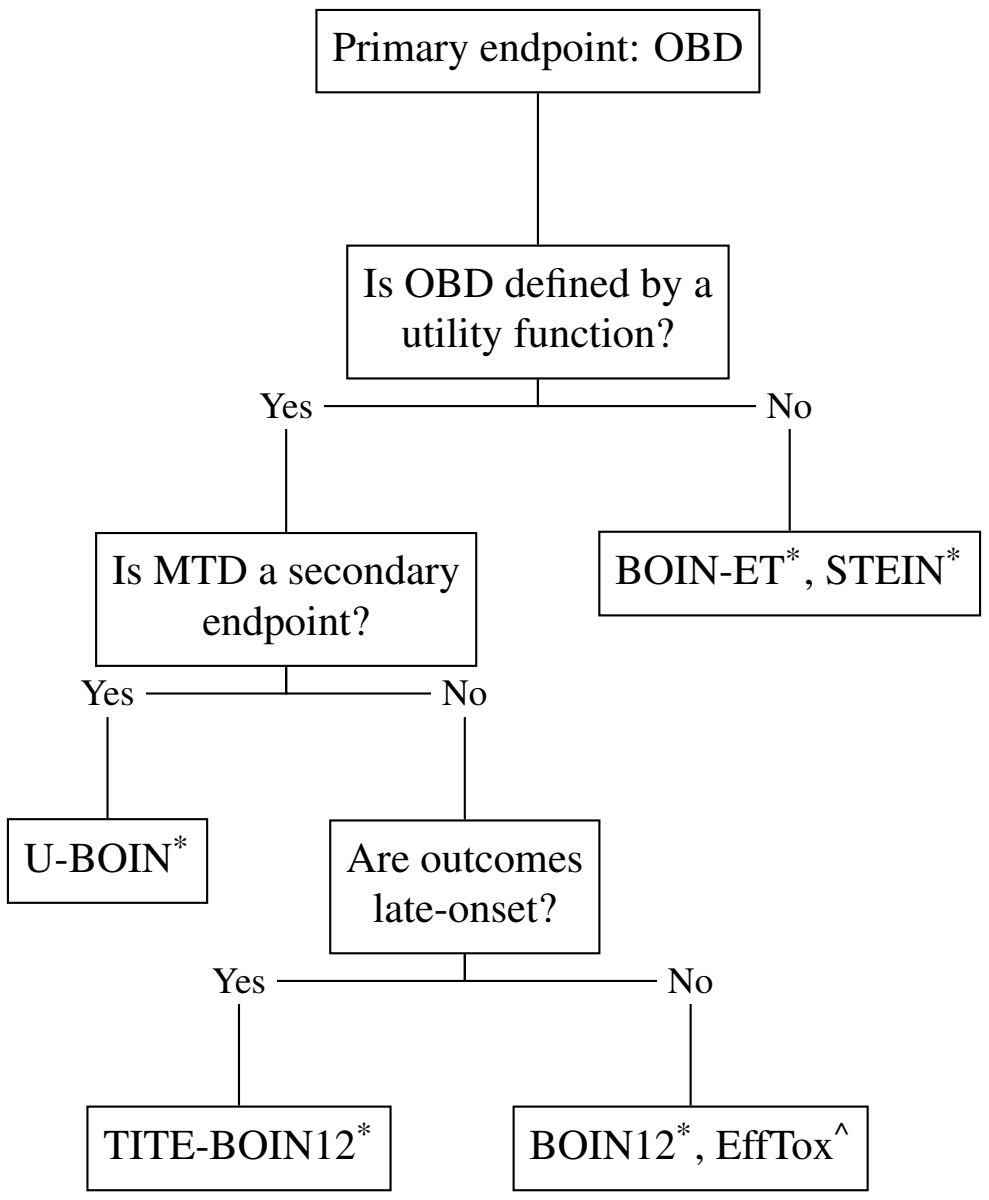
\begin{figure}[H]
  \centering
  \begin{subfigure}{\textwidth}\centering
  \adjustbox{max width=\linewidth, max totalheight=0.4\textheight}{%
    \begin{forest} mdatree
    [Primary endpoint: MTD
      [{Single agent or\\combination therapy?}
        [{With prior\\information?}, EL={Single agent}
          [{iBOIN\ma, iKeyboard\ma,\\iCRM\mb}, EL={Yes}]
          [{Single or multiple/\\continuous toxicity?}, EL={No}
            [{Is toxicity\\late-onset?}, EL={Single}
              [{TITE-BOIN\ma, TITE-Keyboard\ma,\\TITE-CRM\mb}, EL={Yes}]
              [{BOIN\ma, Keyboard\ma, 3+3\ab,\\CRM\mb, BMA-CRM\mb, EWOC\mb, BLRM\mb}, EL={No}]
            ]
            [{MT-BOIN\ma, gBOIN\ma,\\MC-CRM\mb}, EL={Multiple/continuous}]
          ]
        ]
        [{Single MTD or\\MTD contour?}, EL={Combination}
          [{BOIN Comb\ma,\\Keyboard Comb\ma}, EL={Single MTD}]
          [{Waterfall\ma}, EL={MTD contour}]
        ]
      ]
    ]
    \end{forest}}
  \caption{MTD branch (primary endpoint: toxicity).}
  \end{subfigure}

  \vspace{1.5em}

  \begin{subfigure}{\textwidth}\centering
  \adjustbox{max width=\linewidth}{%
    \begin{forest} mdatree
    [Primary endpoint: OBD
      [{Is OBD defined by a\\utility function?}
        [{Is MTD a secondary\\endpoint?}, EL={Yes}
          [{U-BOIN\ma}, EL={Yes}]
          [{Are outcomes\\late-onset?}, EL={No}
            [{TITE-BOIN12\ma}, EL={Yes}]
            [{BOIN12\ma, EffTox\mb}, EL={No}]
          ]
        ]
        [{BOIN-ET\ma, STEIN\ma}, EL={No}]
      ]
    ]
    \end{forest}}
  \caption{OBD branch (primary endpoint: safety and efficacy).}
  \end{subfigure}

  \caption{Full decision workflow for Phase~I dose-finding interventional oncology trial
  design, split by primary endpoint: (a)~MTD branch and (b)~OBD branch.
  \ma model-assisted, \mb model-based, \ab algorithm-based.}
  \label{fig:mdatree-full}
\end{figure}

We walk the decision tree node by node. When a model resolves a node, it is deterministically routed to a deeper node in the tree. The leaf reached is mapped to a design-method class. Please refer to Figure~\ref{fig:mdatree-full}.

\subsection{Node-by-node resolution}
\begin{enumerate}\setlength{\itemsep}{2pt}
  \item \textbf{In scope of dose-finding?} A trial is in scope unless it is not
  dose-finding (fixed-dose or Recommended Phase 2 Dose confirmation, vaccine-immunogenicity, or
  expansion-only) or has no escalation dimension. Out-of-scope trials receive no
  design.
  \item \textbf{MTD or OBD?} A dose-finding trial adopts either a Maximum
  Tolerated Dose (MTD) or an Optimal Biological Dose (OBD) objective \citep{Yuan2024DoseOptimization}.
  Read from the Phase~I objective.
  \item \textbf{Single agent or combination?} Whether adjacent dose levels follow
  a non-monotonic toxicity order. A partially ordered (non-monotonic) dose space
  indicates a combination design, while a monotonic one indicates a single agent.
  \item \textbf{Combination target (combination only).} If the design delivers a
  single recommended dose pair $\rightarrow$ \textbf{BOIN-Comb}. If it delivers
  the MTD \emph{curve} $\rightarrow$ \textbf{Waterfall}. This is a leaf node. The
  single-agent late-onset node is not on the combination path.
  \item \textbf{Informative prior?} For simplicity of design, this study resolves
  this node to negative.
  \item \textbf{Single or multiple/continuous DLT?} Whether the trial monitors
  one binary DLT endpoint, or generalizes to either multiple concurrent toxicity
  types or a graded (non-binary) toxicity outcome.
  \item \textbf{Late onset? (BOIN vs.\ TITE-BOIN)} Whether each patient's
  dose-limiting toxicity is fully observed within a fixed window before the next
  escalation decision  \citep{Yuan2022ModelAssisted}. First, if the escalation gate of a trial opens
  before one cycle of dose-limiting toxicity (DLT) window closes, this trial
  qualifies for late onset. Otherwise, we compute $W/I$, the DLT assessment
  window divided by the interpatient accrual interval, and decide. Full decision
  rules are detailed in \S\ref{apd:lot}.
  \item \textbf{3 levels of OBD branch nodes.} Resolve utility vs.\ independent admissibility,
  MTD-as-secondary, and whether the dose-selection outcomes are late-onset,
  yielding \textbf{BOIN-ET}, \textbf{U-BOIN}, \textbf{BOIN12}/\textbf{EffTox},
  or \textbf{TITE-BOIN12}.
  \item \textbf{Leaf $\rightarrow$ label.} The leaf is a set of admissible
  designs, ranked into Primary/Secondary/Acceptable tiers; the reported label is
  the Primary design.
\end{enumerate}

\subsection{Late-onset determination}
\label{apd:lot}
\begin{enumerate}\setlength{\itemsep}{2pt}\setcounter{enumi}{9}
  \item \textbf{$W$. DLT assessment window (days).} The registry
  primary-outcome time frame when it is the DLT/MTD/safety endpoint. Otherwise
  the protocol DLT-definition clause. Else cycle length $\times$ number of DLT
  cycles. Guards: assessment \emph{cadence} is not the window (``every 3 wk for
  9 wk'' gives $W\!=\!63$, not $21$). A response/survival time frame is not a DLT
  window. If the DLT is defined on a combination begun after a single-agent
  run-in, $W$ starts with the combination. For multi-stage designs $W$ is the
  governing (longest, latest-stage) window.
  \item \textbf{$I$. Interpatient interval, as a range [fast, slow].} The
  \emph{fast} bound is a protocol-mandated stagger if stated ($I\!\ge\!W$ for
  cohort-gated designs). A short ``dose $x$ days apart'' safety floor counts
  here, not as the accrual pace. The \emph{slow} bound is the registry accrual,
  \[
    I_{\text{slow}}=\frac{t_{\text{complete}}-t_{\text{start}}-W}{N_{\text{arm}}-1}, N_{\text{arm}}=N_{\text{total}}/k,
  \]
  where $k$ is the number of parallel dose-finding cohorts. $W$ is subtracted
  because the completion date trails the last enrollment by about one window.

  \item \textbf{Two-bound ratio.} The window-to-interval ratio spans $W/I_{\text{slow}}$ (lower bound, slowest accrual) to $W/I_{\text{fast}}$ (upper bound).

  \item \textbf{Decision (escalation gate is primary).} (i) The protocol opens
  the next dose \emph{before} the window closes $\rightarrow$ \textbf{TITE-BOIN}.
  (ii) Escalation waits for the full window (cohort-complete, sentinel, or
  single-patient full-window) $\rightarrow$ \textbf{BOIN}, regardless of registry
  accrual. Therefore, an expansion cohort that inflates whole-trial accrual cannot force
  a false TITE. (iii) No gate stated $\rightarrow$ fall back to accrual:
  $W/I_{\text{slow}}\ge 4 \rightarrow$ \textbf{TITE-BOIN}, else \textbf{BOIN}.
\end{enumerate}

\subsection{OBD variant}
The OBD branch has two windows, toxicity $W_T$ and the dose-\emph{selection}
efficacy endpoint $W_E$ (binarized ORR, a PD biomarker, or landmark PFS, not
long-term survival). $\max(W_T,W_E)/I \ge$ cutoff $\rightarrow$
\textbf{TITE-BOIN12}, else \textbf{BOIN12}/\textbf{EffTox}.

\section{Model List and Inference Configuration}
\label{apd:models}
All models were accessed through a single unified serving layer (Databricks
Model Serving / Foundation Model APIs) to standardize access and pin versions
across providers. Decoding was deterministic
with identical settings for every model and input configuration. Model details are listed in Table~\ref{tab:models}.

\begin{table}[H]
\centering\small
\caption{Serving endpoints for the five evaluated models.}
\vspace{6pt}
\label{tab:models}
\begin{tabular}{ll}
\toprule
Model & Serving endpoint \\
\midrule
Claude Opus 5      & \texttt{databricks-claude-opus-5} \\
GPT-5              & \texttt{openai-gpt-5} \\
OpenAI o3          & \texttt{openai-gpt-o3} \\
gpt-oss-120b       & \texttt{databricks-gpt-oss-120b} \\
Llama 4 Maverick   & \texttt{databricks-llama-4-maverick} \\
\bottomrule
\end{tabular}
\end{table}

\section{Prompts}
\label{apd:prompts}
Every input configuration shares a system block defining the role, an anti-lookup
instruction, a rule for out-of-scope situation, and the class taxonomy
(Appendix~\ref{app:classes}). The tree input configurations (D, E) additionally include the
decision tree and a node-by-node walk instruction. The per-input-configuration user block
carries the fields of Appendix~\ref{app:conditions}. It is followed by representative excerpts of
the tree-walk system prompt. Full verbatim prompts for every input configuration are
released with the code.

\begin{quote}\footnotesize\itshape
You are a professional clinical trial biostatistician specializing in Phase~I
and Phase~I/II oncology dose-finding trial design. [\dots] Do not attempt to
identify this specific trial or look up related materials. [\dots] Base your
recommendation solely on the information given here [\dots], not on recalled
facts about any specific real trial.

If the trial described is out of scope [\dots] (for example a PK/PD-only,
imaging, single fixed-dose, or vaccine-immunogenicity-only study), mark it out of
scope and do not force a recommendation.

[Tree input configurations] Follow the tree closely [\dots], answering each node you pass
through, in order, with a brief rationale, until you reach a leaf. Record every
node you visit in \texttt{node\_answers}. [\dots] The distinction between
single-agent and combination dose-escalation is whether the dose-LEVEL sequence
has a joint (non-monotonic) toxicity order --- NOT how many agents are
administered or how many agents' doses vary.
\end{quote}

\paragraph{Redaction.} Model-visible inputs exclude the originally adopted
design, reported results, and any design-revealing registry field. The case is
identified by an anonymized case ID only.

\end{document}